\pdfoutput=1

\documentclass[11pt]{article}

\usepackage{acl}

\usepackage{times}
\usepackage{latexsym}

\usepackage[T1]{fontenc}

\usepackage[utf8]{inputenc}

\usepackage{microtype}

\usepackage{inconsolata}

\usepackage{geometry}
\usepackage[utf8]{inputenc} 
\usepackage[T1]{fontenc}   
\usepackage{fullpage}
\usepackage{floatrow}

\usepackage{blindtext}

\date{}

\usepackage{array}
\usepackage{arydshln}
\usepackage{amsmath}
\usepackage{amssymb}
\usepackage{mathtools}
\usepackage{amsthm}

\usepackage{hyperref} 
\usepackage{cleveref}
\newcommand{\comment}[1]{}
\usepackage[T1]{fontenc}    
\usepackage{url}            
\usepackage{booktabs}       
\usepackage{amsfonts}       
\usepackage{nicefrac}       
\usepackage{microtype}      
\usepackage{algorithm}
\usepackage{algorithmic}
\usepackage{algorithmic}
\usepackage{algorithmic}
\usepackage{algorithmic}
\usepackage{lipsum}
\usepackage{enumitem}
\usepackage{amsmath, amssymb, amsthm}
\usepackage{wrapfig}
\crefname{equation}{}{}
\Crefname{equation}{}{}

\usepackage[utf8]{inputenc}
\usepackage{mathtools}
\usepackage{subcaption}
\usepackage{graphicx}
\usepackage[font=small,labelfont=bf]{caption}
\usepackage{bbm}
\usepackage{array}
\usepackage{arydshln}

\usepackage{multirow}

\usepackage{url}            
\usepackage{booktabs}       
\usepackage{amsfonts}       
\usepackage{nicefrac}       
\usepackage{xcolor}
\usepackage{microtype}      
\usepackage{amssymb,amsmath, amsthm}
\usepackage[utf8]{inputenc}
\usepackage{mathtools}
\usepackage{subcaption}
\usepackage{tikz}
\usetikzlibrary{calc}
\usepackage{graphicx}
\usepackage{caption}
\usepackage{bbm}
\usepackage{bm}
\usepackage{array}
\usepackage{arydshln}
\usepackage{xspace}

\usepackage{algorithmic}
\usepackage[colorinlistoftodos,prependcaption]{todonotes}

\definecolor{gr}{rgb}{0.25, 0.25, 0.25}

\newcommand{\rbb}{\mathbb{R}}

\newcommand{\ec}{M}

\newcommand{\ic}{k}

\newcommand{\rmin}{r_{\text{min}}}
\newcommand{\rmax}{r_{\text{max}}}

\newcommand{\lwb}{\mathbf{W}}
\newcommand{\lab}{\mathbf{A}}
\newcommand{\lbb}{\mathbf{B}}
\newcommand{\lsb}{\mathbf{S}}
\newcommand{\labg}{\overline{\mathbf{A}}}
\newcommand{\lbbg}{\overline{\mathbf{B}}}
\newcommand{\lwbg}{\overline{\mathbf{W}}}

\newcommand{\bdat}{\mathcal{B}_\ic}

\newcommand{\defeq}{\vcentcolon=}

\newcommand{\ti}{(t)}

\newcommand{\lr}{\eta}

\newcommand{\hetlo}{\textsc{HetLoRA}\xspace}
\newcommand{\homlo}{\textsc{HomLoRA}\xspace}

\newcolumntype{?}{!{\vrule width 1pt}}

\newcommand{\st}{\mathcal{S}^{\ti}}

\setlist[enumerate]{label=\roman*),
                    leftmargin=2em, 
                    }

\theoremstyle{plain}

\newtheorem*{thm*}{Theorem}

\graphicspath{{Figures/}}

\crefname{equation}{}{}
\Crefname{equation}{}{}
\crefname{thm}{theorem}{theorems}
\Crefname{thm}{Theorem}{Theorems}
\crefname{clm}{claim}{claims}
\Crefname{clm}{Claim}{Claims}
\Crefname{coro}{Corollary}{Corollaries}
\Crefname{lem}{Lemma}{Lemmas}
\Crefname{sec}{Section}{Sections}
\crefname{app}{appendix}{appendices}
\Crefname{app}{Appendix}{Appendices}
\crefname{prop}{proposition}{propositions}
\Crefname{prop}{Proposition}{Propositions}
\Crefname{propty}{Property}{Properties}
\crefname{figure}{fig.}{figures}
\Crefname{figure}{Fig.}{Figures}
\crefname{defn}{definition}{definitions}
\Crefname{defn}{Definition}{Definitions}
\crefname{fact}{fact}{facts}
\Crefname{fact}{Fact}{Facts}
\crefname{appendix}{appendix}{appendices}
\Crefname{appendix}{Appendix}{Appendices}
\crefname{algo}{algorithm}{algorithms}
\Crefname{algo}{Algorithm}{Algorithms}
\crefname{algorithm}{algorithm}{algorithms}
\Crefname{algorithm}{Algorithm}{Algorithms}
\crefname{tbl}{table}{table}
\Crefname{tbl}{Table}{Table}
\crefname{table}{table}{table}
\Crefname{table}{Table}{Table}
\crefname{algorithm}{algorithm}{algorithms}
\Crefname{algorithm}{Algorithm}{Algorithms}

\crefname{conj}{conjecture}{conjectures}
\Crefname{conj}{Conjecture}{Conjectures}
\crefname{obs}{observation}{observations}
\Crefname{obs}{Observation}{Observations}

\title{Heterogeneous LoRA for Federated Fine-tuning of \\On-Device Foundation Models
}

\author{ \bf
Yae Jee Cho$^1$\thanks{*Work done while at Google Research. Corresponding authors: yaejeec@andrew.cmu.edu, luyangliu@google.com}
, Luyang Liu$^2$, Zheng Xu$^2$, Aldi Fahrezi$^2$, Gauri Joshi$^1$ \\
$^1$Carnegie Mellon University, $^2$Google Research\\
\normalsize {\texttt{yaejeec@andrew.cmu.edu, \{luyangliu,xuzheng,aldifahrezi\}@google.com,}}\\ \normalsize{\texttt{gaurij@andrew.cmu.edu}}
}

\begin{document}
\maketitle
\begin{abstract}
Foundation models (FMs) adapt well to specific domains or tasks with fine-tuning, and  federated learning (FL) enables the potential for privacy-preserving fine-tuning of the FMs with on-device local data. 
For federated fine-tuning of FMs, we consider the FMs with small to medium parameter sizes of single digit billion at maximum, referred to as \emph{on-device FMs (ODFMs)} that can be deployed on devices for inference but can only be fine-tuned with parameter efficient methods. 
In our work, we tackle the data and system heterogeneity problem of federated fine-tuning of ODFMs by proposing a novel method using heterogeneous low-rank approximations (LoRAs), namely \hetlo. First, we show that the naive approach of using homogeneous LoRA ranks across devices face a trade-off between overfitting and slow convergence, and thus propose 
\hetlo, which allows \textit{heterogeneous ranks} across client devices and efficiently aggregates and distributes these heterogeneous LoRA modules. By applying rank self-pruning locally and sparsity-weighted aggregation at the server, \hetlo~combines the advantages of high and low-rank LoRAs, which achieves improved convergence speed and final performance compared to homogeneous LoRA. Furthermore, \hetlo offers enhanced computation efficiency compared to full fine-tuning, making it suitable for federated fine-tuning across heterogeneous devices.
\end{abstract}

\section{Introduction} \label{sec:intro} 


The emerging foundation models (FMs)~\cite{rishi2022fm,zhou2023fmsurvey,radford2021clip,devlin2019bert,openai2023gpt4,chowd2022palm,touvron2023llama,brown2020gpt3,chowd2022palm,driess2023palme,google2023palm2} have shown remarkable zero/few shot learning capabilities, performing well on a variety of tasks including text/image generation with prompts, language translation, solving math problems, and conversing in natural language. 
Standard FMs, however, demand costly resources for directly fine-tuning their entire parameter space. To tackle this issue, many recent works have proposed different parameter-efficient fine-tuning (PEFT) methods of FMs such as prompt tuning~\cite{lester2021prompt}, utilizing adapters~\cite{houlsby2019adapter}, or low-rank adaptation (LoRA) of the original model~\cite{hu2021lora} which freezes the original pre-trained parameters of the FM and train additional, smaller number of parameters instead. 

These PEFT methods, however, assume that i) FMs are deployed to and trained with the data of a \textit{single} machine/client for adaptation to the downstream task and that ii) the client has enough resources to even fit a standard FM of hundred billion size for, at least, inference. In practice, there are frequently cases where we are interested in fine-tuning FMs for on-device private data that is distributed across multiple devices (clients). For instance, sensitive and private data such as medical information or law-related documents may be hard to collect centrally in a private manner and fine-tuning of the FMs may need to be done at the edge~\cite{manoel2023fedmed, shoham2023medfedbert,zhang2023fedlegal}. 

In our work, we focus on such federated fine-tuning scenarios, where we train a set of parameters collaboratively across clients to obtain a global set of parameters that can be plugged in to the FM for the targeted downstream task. Note that federated fine-tuning is orthogonal to personalization of FMs in federated learning (FL)~\cite{tao2023fmper}, which aims to train parameters that perform well for individual clients rather than general downstream tasks. We also define \textit{on-device FMs (ODFMs)} as models with few billion parameters at max that are able to fit into memory on limited capacity clients considering current hardwares. 



\setlength\intextsep{0pt}
\begin{table}[!t] \centering 
\setlength\tabcolsep{6pt} \small 
\begin{tabular}{@{}lccc@{}} 
\toprule  
 &  Zero-Shot & Few-Shot  & Full-Training   \\ \hline
PaLM 2 XXS & $2930.23$ &  $2541.86$ & $\mathbf{23.71}$     \\  \hdashline
PaLM 2 XS & $2712.86$ & $481.95$  & $\mathbf{18.32}$     \\ 
 \bottomrule  \label{tab:agisize}
\end{tabular} \vspace{-1.5em}
\caption{Perplexity of PaLM 2 for zero-shot, few-shot (5 communication rounds), and full federated fine-tuning (200 communication rounds) for chat response on the multi-session chat data (further experimental details are in \Cref{sec:exp}.) \vspace{0.3em} }
\end{table}

Federated fine-tuning of ODFMs entails unique challenges non-present in either the standard PEFT of FMs or the standard federated training of models that are not FMs. First, FMs have their zero/few-shot learning capability often supported by their large parameter space that is trained on massive data. However, as we show in Table 1 and also presented by previous literature~\cite{kojima2022fmzeroshot}, FMs' performance deteriorates as their sizes get smaller and federated fine-tuning may not merely be useful but \textit{inevitable} for ODFMs to perform well for downstream tasks on devices. 

Moreover, devices have limited and heterogeneous system capabilities~\cite{wang2019systemc, bonawitz2016practical} and data distributions~\cite{sahu2019federated}. A suitable PEFT method that flexibly adapts to such heterogeneity across devices should be investigated for federated fine-tuning of ODFMs. Previous work evaluated PEFT with FL via performing a general evaluation over different PEFT methods na\"ively combined with FL~\cite{guo2022promptfl,zhang2023pretrainfinetune,chen2022fedtune,wortsman2023distributedlofi,yu2023ffm}. However, they do not consider the practical setting for ODFMs where PEFT methods are catered to the system and data heterogeneity of clients. 

In our work. we focus on one of the most prominent PEFT methods, LoRA~\cite{hu2021lora} which proposes to train low-rank approximations of the original model. Using LoRA, the number of trainable parameters is greatly reduced to at most $0.02\%$ of the original ODFM size (see \Cref{tab:lorasize}). The simplest way to apply LoRA to federated fine-tuning is training with homogeneous rank $r$ across the clients as one would train any global model with FL. However, this does not cater to the heterogeneity in FL, where it is even difficult to choose the right LoRA rank for resource limited mobile devices with natural system and data heterogeneity. %

\begin{figure}[!t] \centering
    \includegraphics[width=1\textwidth]{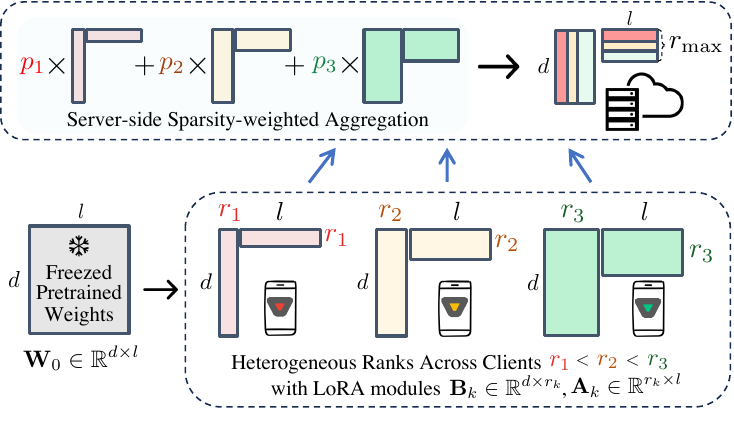} \vspace{-2.5em}
    \caption{\small Overview of heterogeneous rank deployment of LoRA: the pretrained weights $\mathbf{W}_0$ are stored on-device and heterogeneous ranks are assigned to different clients with $r_{\text{min}}=r_1<r_2<r_3=r_{\text{max}}$. In our proposed \hetlo, the server receives the trained heterogeneous LoRA modules and aggregates them with sparsity-weighted aggregation to update the global LoRA module.  \vspace{0.5em} 
    }
\label{fig:setting}
\end{figure}

\begin{table*}[!t]  \centering \small \renewcommand{\arraystretch}{0.8}
\begin{tabular}{l|cccccccc}\\\toprule  
&$r=1$ & $r=5$ & $r=10$ & $r=20$  & $r=50$  & $r=100$  & $r=150$ & $r=200$ \\\midrule
PaLM 2 XXS, PaLM 2 XS & $0.02\%$ & $0.11\%$ & $0.21\%$ & $0.42\%$ & $1.05\%$ & $2.10\%$ & $3.14\%$ & $4.19\%$\\
\bottomrule  \vspace{-1em} 
\end{tabular}
\caption{\small Percentage of the LoRA parameters' size for different ranks $r$ compared to the original pre-trained ODFM's parameter size. Even for large ranks such as $r=200$ the trainable LoRA parameters' size compared to the original pre-trained ODFM size is less than $5\%$ for both PaLM 2-XS and PaLM 2-XXS. \vspace{-1em} }\label{tab:lorasize}
\end{table*}

To this end, we propose heterogeneous LoRA, namely \hetlo in short, for federated fine-tuning to cater to system and data heterogeneity and outperform the na\"ive combination of LoRA and federated fine-tuning where homoegneous ranks are applied across clients. We show the performance of PaLM 2~\cite{google2023palm2} of XXS and XS size for chat responses on the multi-session chat data~\cite{xu2021msc} and text summarization for the Reddit data~\cite{volske2017reddit}, both which are real world data from clients. Our contributions can be summarized as follows: \vspace{-0.3em}
\begin{itemize}[leftmargin=*]
\setlength\itemsep{0.1em}
    \item We propose \hetlo that can apply different rank LoRA modules to different clients to cater to the heterogeneous system capabilities and data complexities of the clients, via utilizing rank self-pruning and sparsity-weighted aggregation.

    \item We show the performance of na\"ively applying LoRA with homogeneous ranks across clients for federated fine-tuning, and show that while large ranks help in speeding-up training, they lead to faster overfitting while smaller ranks are slower in training but does not suffer from overfitting. 
     
    \item We then evaluate \hetlo to show that it outperforms na\"ively applying homogeneous ranks across clients in terms of both training speed, communication/computation efficiency, and final performance, gaining the best of both worlds of homogeneous LoRA with high and low ranks. 
    
\end{itemize} 



\section{Related Work}\vspace{-0.5em}
\textbf{Parameter-Efficient Fine Tuning.} There has been a plethora of recent work on PEFT which either trains a subset of parameters within the existing FM whilst other parameters are freezed or introduces an additional set of trainable parameters whilst keeping the original FM freezed. For the former, methods such as head or bias fine-tuning~\cite{wei2021pretrained,bu2022differentially,lee2019would,zaken2021bitfit} has been explored, while for the latter, methods such as adapters~\cite{houlsby2019adapter}, prompt~\cite{lester2021prompt} or prefix-tuning~\cite{li2021prefix}, and low-rank approximation~\cite{hu2021lora} has been proposed. While these number of methods has been proven to perform as well as full model fine-tuning with just a few number of parameters for the centralized setting, it has not been thoroughly explored how these methods perform for a much smaller FM such as ODFMs, in the decentralized setting where clients' system-capacities can be heterogeneous and much limited. 

\textbf{Federated Fine-Tuning.} Recently, interest in the intersection of FMs and FL has notably increased~\cite{zhou2023fmsurvey,yu2023ffm}. Many recent work has proposed to combine the PEFT methods devised for the centralized setting to FL such as training prompts or adapters collaboratively with FL~\cite{guo2022promptfl,chen2022fedtune, jianyi2022fedgpt,shy2023peftfit,gwen2023lastlayerfm}. Another line of work has proposed to perform a few-shot or nearly zero-shot training of FMs with FL for improved communication-efficiency~\cite{wortsman2023distributedlofi,zhang2023pretrainfinetune}. However, these work either overlooks that most devices do not have the resource to fit a general FM~\cite{touvron2023llama,brown2020gpt3} (>8B parameters) even for inference or does not consider the heterogeneous system capacities of the clients. It is detrimental to consider these factors since FMs that actually fits to the devices in FL are much smaller, making them weaker in the general intelligence capabilities, and also heterogeneous system capacities may prohibit deploying same sized PEFT parameters across clients.

Only a few number of recent work has looked in to using LoRA for FL. For instance, in~\cite{baba2023slora}, the importance of the initialization for the LoRA modules is evaluated where they propose to train the LoRA modules with FL and then perform singular value decomposition (SVD) to gain a good initialization of the LoRA modules. However, the training process of LoRA itself is not altered to adapt to heterogeneous system capabilieis of devices. Another recent work~\cite{yi2023fedlora} has evaluated LoRA in the context of personalized FL, but other than applying LoRA to personalization, the LoRA method itself is, again, not changed. Our work proposes heterogeneous LoRA for federated fine-tuning where heterogeneous ranks are deployed and trained across clients by a new algorithm that includes rank self-pruning and sparsity weighted aggregation. 


\section{Federated Fine-Tuning with LoRA} \label{sec:pf}
\subsection{Preliminaries}
Formally, we define the pre-trained ODFM as $\lwb_0\in\rbb^{d\times l}$ and the trainable low-rank decomposed matrix as $\Delta\lwb\in\rbb^{d\times l}$. In standard LoRA~\cite{hu2021lora} under the centralized setting, the low-rank decomposition of $\Delta\lwb$ is constructed such that  $\Delta\lwb=\lbb\lab$ where $\lbb\in\rbb^{d\times r}$ and $\lab\in\rbb^{r\times l}$ are the low rank decomposition of $\Delta\lwb$ with identical rank $r$. Now, let us consider LoRA for federated fine-tuning where there are $M$ total clients. Each client $\ic\in[\ec]$ has private data $\bdat$ and its corresponding local empirical loss function $F_\ic(\lwb)=\frac{1}{|\bdat|}\sum_{\xi \in \bdat} \ell(\lwb,\xi)$, where $\ell(\lwb,\xi)$ is the loss for model $\lwb$ at data sample $\xi$. The optimization task for federated fine-tuning is to collaboratively find the global parameters which we define as $\lbbg$ and $\labg$, given the pretrained knowledge $\lwb_0$ that can minimize the global objective $F(\lwbg)=\frac{1}{M}\sum_{\ic=1}^{\ec}F_\ic(\lwbg)$ where $\lwbg=\lwb_0+\lbbg~\labg$. 
Later in the paper, when introducing heterogeneous LoRA we truncate the LoRA modules' rank dimension, for example from $\lbb\in\rbb^{d\times r},~\lab\in\rbb^{r\times l}$ to $\lbb'\in\rbb^{d\times r'},~\lab'\in\rbb^{r'\times l}$ where $r'<r$.
Throughout the paper, we denote such truncation of a matrix with the $:$ symbol for each row and column at the subscript. For instance, for truncation to $r'<r$ at the column for the matrix $\lbb\in\rbb^{d\times r}$, we keep all the columns until $r'$ and omit the last $r-r'$ columns and denote the resulting matrix it as $\lbb_{:,:r'}$.

\subsection{Na\"ive Case: Homogeneous LoRA} \vspace{-0.2em} \label{subsec:homolora}

A straightforward way to perform federated fine-tuning with LoRA is to train the LoRA modules $\lbb,~\lab$ with homogeneous rank $r$ across all clients with standard FL~\cite{mcmahan2017communication}. Specifically, first the clients have the pre-trained ODFM weights $\lwb_0$ stored in their devices prior to training for the forward pass when training the LoRA modules. Then, the server sends the global LoRA modules $\lbbg^{(t)},~\labg^{(t)}$ to the set of $m$ selected clients $\mathcal{S}^{(t)}$ per communication round $t$. 
Each selected client $k\in\mathcal{S}^{(t)}$ trains the LoRA modules on their local data for a few local iterations (usually with mini-batch SGD) and send the updated modules $\lbb_k^{(t)},~\lab_k^{(t)}$ back to the server. The server then updates the global LoRA modules accordingly to $\lbbg^{(t+1)}=\sum_{k\in\mathcal{S}^{(t)}}\lbb_k^{(t)}/m,~\labg^{(t+1)}=\sum_{k\in\mathcal{S}^{(t)}}\lab_k^{(t)}/m$ and sends back to the next set of selected clients for the next communication round. This training process is nearly identical to the standard FL algorithm~\cite{mcmahan2017communication} except that the pretrained weights $\lwb_0$ are freezed and locally stored in the clients' devices and only the LoRA moduels are trained and communicated.

Instead of such homogeneous rank deployment across all clients, it is not only possible but more practical to use heterogeneous rank deployment for federated fine-tuning. This involves training LoRA modules with varying ranks on different clients, based on their system capabilities. Such setting is motivated and often required from the system constraints of the clients~\cite{wang2021field} where most of the clients are only capable of having smaller ranks while a few can handle larger ranks. However, this approach poses challenges in aggregating and redistributing the LoRA modules. To address these challenges, we introduce a solution called \hetlo, which pushes the limits beyond homogeneous LoRA deployment.

\begin{figure*}[!t] \centering
\includegraphics[width=1\textwidth]{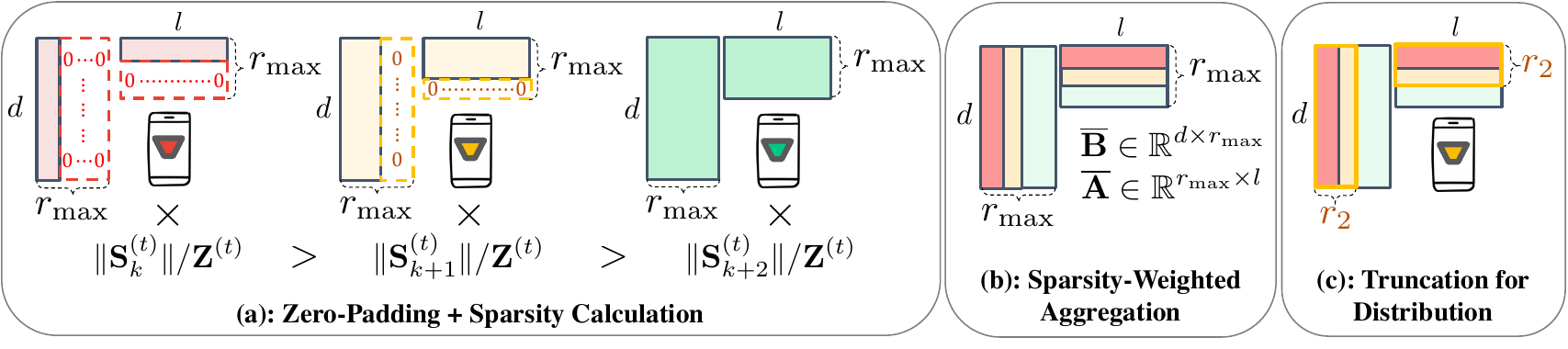} 
\vspace{-2em} 
\caption{\small Overview of the zero-padding, sparsity-weighted aggregation, and truncation method for \hetlo; (a): Zero-pad LoRA modules with smaller ranks to $r_{\text{max}}$ (clients with rank $r_{\text{max}}$ does not need padding) and calculate their sparsity by calculating the Frobenius norm of the reconstructed model $\Delta\lwb_k^{(t)}=\lbb_k^{(t)}\lab_k^{(t)}$; (b): After padding, aggregate all of the clients' LoRA modules with the weights $\|\mathbf{S}_{k}^{(t)}\|/\mathbf{Z}^{(t)}$ calculated by $\Delta\lwb_k^{(t)}$ to get the global LoRA modules; (c): Truncate the global LoRA modules for the specific rank of the next selected client (example for client with rank $r_2$).\vspace{-1em}\label{fig:app1}}
\end{figure*}

\subsection{Proposed Method: Heterogeneous LoRA}  \label{subsec:heterolora}

\textbf{Overview.} Our proposed heterogeneous LoRA method, namely \hetlo, is not restricted to any specific method to assign the ranks to the clients and the clients can decide their respective ranks themselves. For formality, in our paper, we formulate that each client has a rank denoted as $r_k$, within a range of $r_k\in[r_{\text{min}},~r_{\text{max}}],~\forall k$ (see \Cref{fig:setting}). 
\hetlo comprises three steps:
1) Distribution via Truncation, 2) Local Training with Rank Self-Pruning, and 3) Sparsity-Weighted Aggregation of the LoRA modules. These steps are detailed further in the subsequent paragraphs.
An overview of \hetlo is illustrated in \Cref{fig:app1}.

\textbf{1) Distribution via Truncation.} At the begining of each communication round $t$, the server holds initial global LoRA modules $\lbbg^{(t)},~\labg^{(t)}$ with a global rank $r^{(t)}$. The value of the global rank $r^{(t)}$ depends on how we aggregate the heterogeneous rank LoRA modules which is elaborated on in step 3). The server then distributes these global LoRA modules to a subset of selected set of clients $\mathcal{S}^{(t)}$ with heterogeneous ranks $r_k^{(t)},k\in\mathcal{S}^{(t)}$ for local training\footnote{There is a superscript $t$ for the ranks $r_k^{(t)}$ across clients which indicates that in \hetlo~these heterogeneous ranks can be changed over the communication rounds via self-pruning explained in step 2).}.  With the given global LoRA modules, we consider a simple and intuitive method of \textit{truncation} where the server sends $\lbbg^{(t)}_{:,:r_k},~\labg^{(t)}_{:r_k,:}$ to each client $k$ with rank $r_k^{(t)}$ for local training where we omitted the superscript for $r_k$ for simplicity. 

\textbf{2) Local Training with Rank Self-Pruning.} After receiving LoRA modules from the server as $\lbb_k^{(t,0)}=\lbbg^{(t)}_{:,:r_k},~\lab_k^{(t,0)}=\labg^{(t)}_{:r_k,:}$, each client $k\in\mathcal{S}^{(t)}$ performs $\tau$ local iterations of mini-batch SGD on their local data to minimize the local objective $\frac{1}{|\bdat|}\sum_{\xi \in \bdat} \ell((\lbb_k,~\lab_k),\xi | \lwb_0)$, and sends back the updated LoRA modules $\lbb_k^{(t,\tau)}\in\rbb^{d\times r_k^{(t)}}$ and $\lab_k^{(t,\tau)}\in\rbb^{r_k^{(t)}\times l}$ to the server. This is the same process as the standard local training step in vanilla FedAvg~\cite{mcmahan2017communication}. However, we improve this vanilla local training step by adding a rank self-pruning mechanism where clients self-prune their respective ranks depending on the magnitude of the model parameters.

Specifically, we add a regularization term to the original local objective to get $\min_{\lbb_k,~\lab_k}\frac{1}{|\bdat|}\sum_{\xi \in \bdat} \ell((\lbb_k,~\lab_k),\xi | \lwb_0)+\lambda \left\|\lbb_{k,:,r_k\gamma:r_k}\right\|\left\|\lab_{k,r_k\gamma:r_k,:}\right\|$ 
where $\gamma<1$ is a decay-factor that determines how aggressively we want to prune the ranks to a smaller value. The regularization term aims to minimize the norm of the last few ranks, which will become smaller if the first loss term $\frac{1}{|\bdat|}\sum_{\xi \in \bdat} \ell((\lbb_k,~\lab_k),\xi | \lwb_0)$ is not very large. After training with the new local objective we compare the norm of the updated LoRA modules' last few layers $\left\|\lbb_{k,:,r_k\gamma:r_k}\right\|\left\|\lab_{k,r_k\gamma:r_k,:}\right\|$ with the ones from the initially received LoRA modules. If the former is smaller we prune the last few layers (pruning intensity is determined by $\gamma$) and send back the LoRA modules with a smaller rank. This means that for the LoRA modules which incurs a small local loss, i.e., well-trained on the clients' local data, the LoRA modules are more likely to be pruned to a smaller rank. 

Such pruning allows \hetlo to reduce the noise in the LoRA modules introduced by clients having a larger rank than the actual rank that their data complexity requires, and also reduces the complexity of the LoRA modules to improve generalization and prevent overfitting (see \Cref{tab:decay}). Once the rank is pruned for a client, the client saves the updated rank and uses it as the starting rank if selected for future communication rounds. The client then sends back their updated and possibly rank-pruned LoRA modules to the server for the modules to be processed in the next server-side aggregation step.

\textbf{3) Sparsity-Weighted Aggregation. } Finally, the last step of \hetlo is aggregating the received heterogeneous LoRA modules $\lbb_k^{(t,\tau)},~\lab_k^{(t,\tau)},k\in\st$. A straightforward way to aggregate the hetergeneous LoRA modules is using \textit{zero-padding} to all the received LoRA modules with $r_i^{(t)}<\max\{r_k^{(t)}|k\in\st\}$ and then perform simple averaging over the modules. However, such naive aggregation can lead to biasing the model towards higher rank clients even when these clients may not hold valuable training information, i.e., having low data complexity, giving noisy updates. 

In an ideal scenario where we can deploy any rank to any client, deploying higher ranks to the clients with higher data complexity or larger local datasets can retrieve more informative and less sparse updates from the clients. Conversely if we assign higher ranks to the clients whose data complexity is low, the actual rank of the full model from the reconstructed LoRA modules can be smaller than the assigned rank. Thus the higher rank client's update may be unnecessarily over-emphasized in the naive zero padding method. 

Based on this insight we propose a sparsity-weighted aggregation scheme where the server reconstructs these LoRA modules to the full model as $\Delta\lwb_k^{(t)}=\lbb_k^{(t)}\lab_k^{(t)}$ and gets the norm of the singular value vectors from the full models denoted as $\lsb_k^{(t)}$ by calculating $\|\Delta\lwb_k^{(t)}\|_{F}$. Note that the costly process of performing SVD for each of the full model $\Delta\lwb_k^{(t)}$ can be avoided by simply calculating the Frobenius norm of $\Delta\lwb_k^{(t)}$ (see Lemma 1.2 in \cite{guru2012lecnote}). The server then weighs the LoRA modules with aggregation weight $p_k^{(t)}$ which is proportional to the norm of the singular value vectors. Formally, we have the the global LoRA modules updated as $\lbbg^{(t+1)}=\sum_{k\in\mathcal{S}^{(t)}}p_k^{(t)}\lbb_k^{(t)},~\labg^{(t+1)}=\sum_{k\in\mathcal{S}^{(t)}}p_k^{(t)}\lab_k^{(t)}$ where $p_k^{(t)}\defeq\|\lsb_k^{(t)}\|/\mathbf{Z}^{(t)}$ with normalizing factor $\mathbf{Z}^{(t)}\defeq \sum_{k'\in\mathcal{S}^{(t)}}\|\lsb_{k'}^{(t)}\|$. This way, we can de-emphasize the larger rank assigned-clients that have rather less informative updates, and more emphasize the smaller rank assigned-clients that have more informative ones. 




\subsection{Why not Simply Reconstruct First, then Redistribute the LoRA modules?} \label{subsec:recon}
One might ask why not simply reconstruct each of the LoRA modules to the full matrix and aggregate them. Here we show that reconstructing the LoRA modules and aggregating them to get the full model results in a different full model compared to when we aggregate the LoRA modules first and then reconstruct the final model. In \Cref{sec:exp} we also empirically show that reconstructing the LoRA modules to the full model and redistributing them after truncated SVD to the corresponding rank of the clients results in an underwhelming performance compared to \hetlo. 

Let us consider a simple case where there are 2 clients with heterogeneous rank lora modules $\lbb_1\in\rbb^{d\times 1},\lab_1\in\rbb^{1\times l}$ and $\lbb_2\in\rbb^{d\times 1},\lab_2\in\rbb^{2\times l}$ respectively for client 1 and client 2 where the former has rank $1$ and latter has rank $2$. We set the notation for the LoRA modules' $i^{th}$ row and $j^{th}$ column value for $\lbb_k$ and $\lab_k$ as $b_{k,ij}$ and $a_{k,ij}$ respectively. Then with $d=3,~l=2$,  
\comment{
\begin{align}
\lbb_1=\begin{bmatrix}
b_{1,00}  \\
b_{1,10} \\ 
b_{1,20} 
\end{bmatrix},~
\lab_1=\begin{bmatrix}
a_{1,00} & a_{1,01} \\
\end{bmatrix}\\
\lbb_2=\begin{bmatrix}
b_{2,00} & b_{2,01} \\
b_{2,10} &  b_{2,11} \\ 
b_{2,20} &  b_{2,21}
\end{bmatrix},~
\lab_2=\begin{bmatrix}
a_{2,00} & a_{2,01} \\
a_{2,10} &  a_{2,11}  
\end{bmatrix}
\end{align}}
when we reconstruct each of the LoRA modules first and then aggregate the full model we have its $i^{th}$ row and $j^{th}$ column as $(\sum_{k=1}^2b_{k,i0}a_{k,0j})+b_{2,i1}a_{2,1j}$ and aggregating the LoRA modules first and then reconstructing the model has the full model's $i^{th}$ row and $j^{th}$ column as $(\sum_{k=1}^2b_{k,i0})(\sum_{k=1}^2a_{k,0j})+b_{2,i1}a_{2,1j}$.
\comment{
\begin{align*}
\delta (\lbb_1\lab_1+\lbb_2\lab_2)\\
=\begin{bmatrix}
\sum_{i=1}^2b_{i,00}a_{i,00}+b_{2,01}a_{2,10} & \sum_{i=1}^2b_{i,00}a_{i,01}+b_{2,01}a_{2,11} \\
\sum_{i=1}^2b_{i,10}a_{i,00}+b_{2,11}a_{2,10} & \sum_{i=1}^2b_{i,10}a_{i,01}+b_{2,11}a_{2,11} \\
\sum_{i=1}^2b_{i,20}a_{i,00}+b_{2,21}a_{2,10} & \sum_{i=1}^2b_{i,20}a_{i,01}+b_{2,21}a_{2,11} 
\end{bmatrix}
\end{align*}

On the other hand, if we zero-pad the lower rank LoRA modules and aggregate the LoRA modules first and then reconstruct the full model from the LoRA modules we get
\begin{align*}
\begin{bmatrix}
(\sum_{i=1}^2b_{i,00})(\sum_{i=1}^2a_{i,00})+b_{2,01}a_{2,10} & (\sum_{i=1}^2b_{i,00})(\sum_{i=1}^2a_{i,01})+b_{2,01}a_{2,11} \\
(\sum_{i=1}^2b_{i,10})(\sum_{i=1}^2a_{i,00})+b_{2,11}a_{2,10} & (\sum_{i=1}^2b_{i,10})(\sum_{i=1}^2a_{i,01})+b_{2,11}a_{2,11} \\
(\sum_{i=1}^2b_{i,20})(\sum_{i=1}^2a_{i,00})+b_{2,21}a_{2,10} & (\sum_{i=1}^2b_{i,20})(\sum_{i=1}^2a_{i,01})+b_{2,21}a_{2,11} 
\end{bmatrix}
\end{align*}
}

One can observe that the difference between the two models are the cross-terms between the left and right module of different client 1 and 2, i.e., $b_{1,i0}a_{2,0j}+b_{2,i0}a_{1,0j}$ for the $i^{th}$ row and $j^{th}$ column. In other words, when we reconstruct the LoRA modules first and then aggregate them to get the full model, each term in the full model are cross-products between the left and right module of each client and not the cross-products between clients. Thus, reconstructing the LoRA modules loses information on the cross-relation across clients, only retaining the knowledge on the cross-relation between the LoRA modules $\lbb$ and $\lab$. Such observation is also corroborated by the \textit{reconstruction first}'s underwhelming performance in \Cref{tab:final}. \vspace{-.5em}

\begin{table*}[!t] \centering 
\small \caption{Final RougeL score for Reddit text summarization and perplexity for multi-session chat for different federated fine-tuning methods. The blue text indicates the ratio of trained number of parameters compared to the full fine-tuning case. \hetlo outperforms both \homlo and Recon+SVD method, but slightly underperforms the full fine-tuning case. However, compared to full fine-tuning the number of trained parameter is significantly smaller.  \label{tab:final}  \vspace{-1em}
}
\begin{tabular}{@{}lcccc@{}}\toprule
&\multicolumn{2}{c}{Reddit (RougeL)} & \multicolumn{2}{c}{Multi-Session Chat (Perplexity)}\\ \hline
& PaLM 2-XXS & PaLM 2-XS & PaLM 2-XXS & PaLM 2-XS  \\\hline
$\text{Full}$ & $\mathbf{94.56} (\pm 0.01)$ & $\mathbf{94.87} (\pm 0.04)$ & $\mathbf{32.70} (\pm 0.17)$  & $\mathbf{23.40} (\pm 0.36)$ \\ \hline 
$\homlo~r=5$ & $92.57 (\pm 1.56), \textcolor{blue}{\times 0.001}$ & $92.89 (\pm 0.96)$  & $80.51 (\pm 8.32),\textcolor{blue}{\times 0.001}$ &  $64.59 (\pm 9.31)$ \\ \hline 
$\homlo~r=50$ & $70.57 (\pm 2.13), \textcolor{blue}{\times0.01}$ & $84.95 (\pm 1.59)$ & $307.96 (\pm 11.43), \textcolor{blue}{\times 0.01}$ & 
 $167.46 (\pm 1.72)$\\ \hline 
$\text{Recon+SVD}$ & $63.28 (\pm 1.92), \textcolor{blue}{\times 0.003}$ &  $75.17 (\pm 1.25)$ & $323.89 (\pm 20.57), \textcolor{blue}{\times 0.002}$ & $215.63 (\pm 15.38)$ \\ \hline 
$\hetlo~\gamma = 0.99$ & $\mathbf{94.23} (\pm 0.03), \textcolor{blue}{\times 0.003}$ & $\mathbf{94.41} (\pm 0.05)$ & $\mathbf{53.93} (\pm 1.57), \textcolor{blue}{\times 0.002}$ &  $\mathbf{38.76} (\pm 0.52)$\\ 
\bottomrule
\end{tabular}
\end{table*}

\section{Experiments} \vspace{-0.5em} \label{sec:exp}
In this section, we present results for \hetlo and its baselines in terms of the performance on training speed, computation/communication efficiency, and final achieved performance. First, we show the performance of homogeneous LoRA to show how LoRA in general performs for low and high rank values. Second, we demonstrate \hetlo's performance for different $\rmin$ and $\rmax$ values comparing them with full fine-tuning, homogeneous LoRA, and the reconstruction-first method elaborated in \Cref{subsec:recon}. We also conduct an ablation study on \hetlo with varying decay factor $\gamma$ for the rank self-pruning step.
The rank distribution across clients for \hetlo, unless mentioned otherwise, is set to a truncated power-law distribution with $\alpha=0.1$ in the range between $[r_{\text{min}},r_{\text{max}}]$ (inclusively), where the small $\alpha$ value makes the distribution skewed towards smaller ranks. 
All experiments were ran with 3 different random seeds and their average is shown along with the standard deviation.

\comment{
<Outline for Experiment Results - Task: MSC, Reddit, (MedMCQA)>\\
1. Results for Zero-Shot, Few-Shot, full fine-Tuning (Table 1) \\
2. Baseline Results for Homogeneous LoRA (Figure 3) - Lower Bound (rmin) Upper Bound (rmax)\\
3. Main Results for our proposed Heterogeneous LoRA method\\
   - 3.a) Comparison with Homogeneous LoRA, full fine-tuning, and Reconstruction first+SVD\\
   - 3.b) Comparison when no rank decay/regularization is used\\
   - 3.c) Comparison when simple averaging is used
   (overall --> computation/communication cost analysis should be included)\\
4. Additional Results\\
   - 4.a) Fairness across clients' personalized results\\
   - 4.b) What happens when we fix the left side?
 }

\textbf{Model.} We use the Transformer-based language model PaLM 2~\cite{google2023palm2} of size XXS and XS for our experiments which are lightweight enough to fit in to the category of ODFMs~\cite{google_palm2} compared to standard FMs. The LoRA modules are applied to only the self-attention layers as proposed in the original LoRA paper~\cite{hu2021lora}, and their relative number of parameters compared to the original model are shown in \Cref{tab:lorasize}. 

\textbf{Tasks.} The tasks we consider are the chat dialogue from the multi-session chat (MSC) dataset~\cite{xu2021msc} and the text summarization task from the Reddit dataset~\cite{volske2017reddit}. The MSC data is a collection of human-human interactions comprising numerous extended chat sessions, and we use perplexity~\cite{zhang2018perplexity} as the metric which has been used to show the quality of chat responses from generative models from previous literature~\cite{sedoc2019chateval}. We sample 100 users uniformly at random and partition their data for training and evaluation by each \texttt{previous\_dialogs} and \texttt{dialog}. The Reddit text summarization data consists of real users' reddit posts and their summarization, and we use RougeL~\cite{lin2004rouge} as the metric. We use 298 users from Reddit that have at least 100 data samples as the training clients and use another 100 users with at least 100 data samples for evaluation.

\textbf{Local Training.} We use mini-batch size $8$ and number of local iterations $\tau=5$ with the feature length set to $1024$. For the learning rate we perform grid search in $\lr=\{0.1,~0.01,~0.001,~0.0001\}$.  For each MSC and Reddit task, we select 5 and 10 clients per communication round respectively. \vspace{-0.5em}

\begin{figure}[!t]
\centering
\begin{subfigure}{0.49\textwidth}
\centering
\includegraphics[width=1\textwidth]{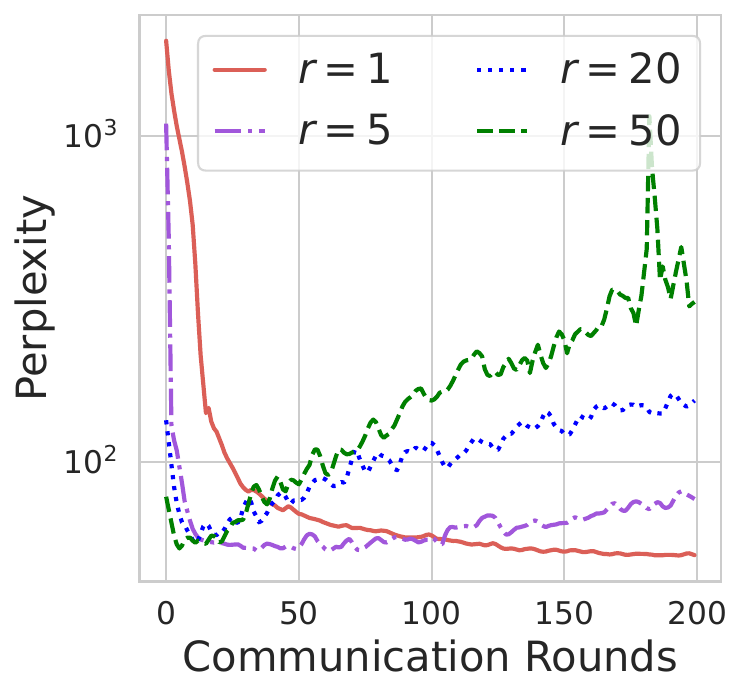}\vspace{-0.5em} \caption{PaLM 2-XXS}\vspace{-1em}
\end{subfigure} 
\centering
\begin{subfigure}{0.49\textwidth}
\centering
\includegraphics[width=1\textwidth]{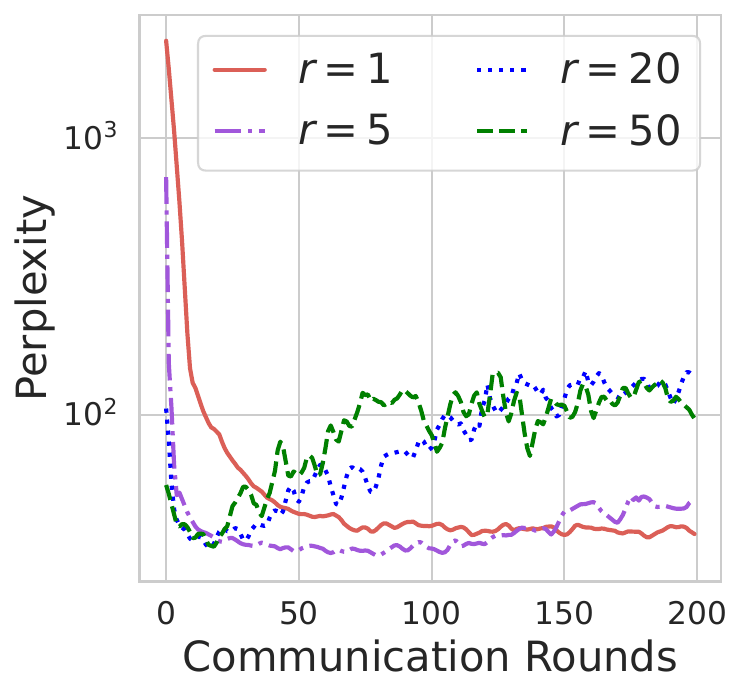}\vspace{-0.5em} \caption{PaLM 2-XS}\vspace{-1em}
\end{subfigure} 
    \caption{\small Performance of homogeneous LoRA for different rank $r$. Higher ranks achieve better performance with fewer communication rounds than the lower ranks, but they overfit quickly. Conversely, the lowest rank $r=1$ achieves low perplexity slower than higher ranks, but without overfitting.\vspace{0.5em}}
\label{fig:homolora} 
\end{figure}

\subsection{Experiment Results}
\vspace{-0.5em}
\textbf{Homogeneous LoRA and the Effect of Ranks $r$.} 
First, we evaluate the performance of federated fine-tuning of the LoRA modules with homogeneous LoRA deployment across clients in \Cref{fig:homolora} for different ranks $r\in[1,5,20,50]$. We observe that a higher rank $r$ for homogeneous LoRA achieves better perplexity floor with fewer communication rounds than the lower ranks but quickly overfits resulting in worse performance compared to the lower ranks after more communication rounds. On the other hand, while the lower rank cases need more communication rounds to achieve good performance, it does not have the problem of overfitting as the higher ranks. 
Hence for homogeneous LoRA, there is a trade-off to consider between low and high ranks, in terms of faster performance achievement and overfitting. Note that these observations are consistent with previous literature in the centralized setting where a higher rank does not necessarily yields the best performance~\cite{hu2021lora,zhang2023adalora}. Next, we show that \hetlo achieves good performance quickly without this overfitting issue, showing better performance than the homogeneous LoRA case.
\comment{
\begin{figure*}[!t]
\centering
\begin{subfigure}{0.245\textwidth}
\centering
\includegraphics[width=1\textwidth]{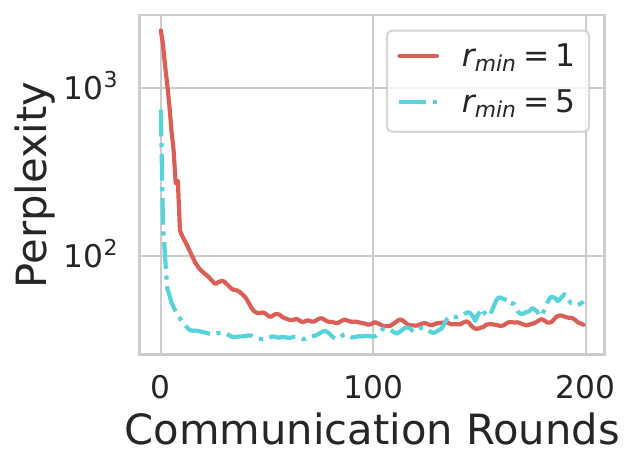}\vspace{-0.5em} \caption{\small $r_{\text{max}}=10$}
\end{subfigure} \hfill
\begin{subfigure}{0.245\textwidth}
\centering
\includegraphics[width=1\textwidth]{fig/hetlora_minmax10_perplexity.pdf}\vspace{-0.5em} \caption{\small $r_{\text{max}}=30$}
\end{subfigure} \hfill
\begin{subfigure}{0.245\textwidth}
\centering
\includegraphics[width=1\textwidth]{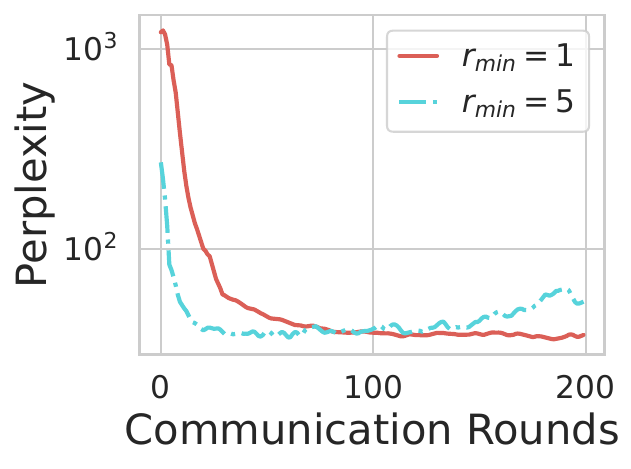}\vspace{-0.5em} \caption{\small $r_{\text{max}}=50$}
\end{subfigure} 
\begin{subfigure}{0.245\textwidth}
\centering
\includegraphics[width=1\textwidth]{fig/hetlora_minmax50_perplexity.pdf}\vspace{-0.5em} \caption{\small $r_{\text{max}}=150$}
\end{subfigure} \vspace{-1em}
 \caption{\small Performance of heterogeneous LoRA with different minimum rank values i.e., $r_{\text{min}}\in[1,5]$. As similar to homogeneous LoRA, larger $r_{\text{min}}$ leads to more overfitting for heterogeneous LoRA, but it is not as severe as the homogeneous LoRA even for larger maximum ranks $r_{\text{max}}=150$ showing that the smaller rank LoRA modules act as a regularizer in heterogeneous LoRA.\vspace{0.3em}} \label{fig:hetlora} 
\end{figure*}}

\begin{figure}[!t]
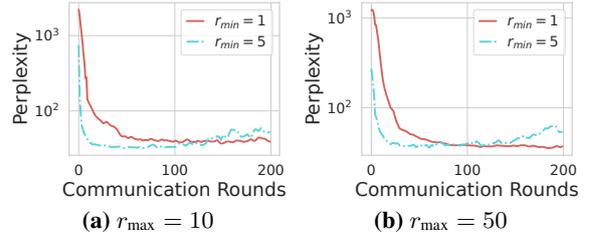

\begin{subfigure}{0.49\textwidth}
    \includegraphics[width=1\textwidth]{fig/hetlora_minmax10_perplexity.pdf} \vspace{-1.7em}\caption{$r_{\text{max}}=10$}\vspace{-1em}
\end{subfigure}
\begin{subfigure}{0.49\textwidth}
    \includegraphics[width=1\textwidth]{fig/hetlora_minmax50_perplexity.pdf} \vspace{-1.7em}\caption{$r_{\text{max}}=50$}\vspace{-1em}
\end{subfigure}
    \caption{\small Performance of \hetlo without rank pruning or and with simple average aggregation. Similar to homogeneous LoRA, larger $r_{\text{min}}$ leads to overfitting for heterogeneous LoRA, but it is not as severe as homogeneous LoRA even for larger maximum rank $r_{\text{max}}=50$ showing that the smaller rank LoRA modules act as a regularizer for \hetlo.\vspace{0.5em}}
\label{fig:hetlora} 
\end{figure}

\textbf{Na\"ive Heterogeneous LoRA and the Effect of $r_{\text{min}}$ and $r_{\text{max}}$.} First, we show the performance of \textit{na\"ive} heterogeneous LoRA \textit{without} self rank-pruning and with only average aggregation instead of the sparsity-weighted aggregation in \Cref{fig:hetlora}. We can see similar observations to those from homogeneous LoRA where a smaller minimum rank $r_{\text{min}}=1$ leads to slower training but better performance while a larger maximum rank leads to faster training but worse performance. However, compared to homogeneous LoRA the overfitting does not get as severe for heterogeneous LoRA even with much larger ranks such as $r_{\text{max}}=50$. We can imply from this result that the smaller rank LoRA modules act as a regularizer in heterogeneous LoRA. Next, we show that by adding the self rank-pruning and sparsity-weighted aggregation, even with $r_{\text{min}}=5$ we are able to prevent overfitting issues and achieve better training speed and final performance than other baselines.


\begin{table*}[!t] \centering \renewcommand{\arraystretch}{0.95}
\small 
\caption{Ablation study on the effect of the decaying factor $\gamma$ for \hetlo's self-rank pruning in the local training step. While aggressive pruning can be harmful to \hetlo's performance, pruning ($\gamma=0.99$) can outperform the case when there is no pruning at all ($\gamma=1$) by reducing the noise introduced by large rank clients with low data complexity.\label{tab:decay}\vspace{-1.5em}}
\begin{tabular}{@{}lcccc@{}}\toprule
&\multicolumn{2}{c}{Reddit (RougeL)} &\multicolumn{2}{c}{Multi-Session Chat (Perplexity)} \\ \hline
& PaLM 2-XXS & PaLM 2-XS & PaLM 2-XXS & PaLM 2-XS  \\\hline
\hetlo, $\gamma = 1$ & $92.17~(\pm 0.08)$  & $91.95~(\pm 0.03)$&  $55.07~(\pm 0.81)$  &  $40.92~(\pm 0.58)$  \\  \hline
\hetlo, $\gamma = 0.99$ & $\mathbf{94.23}~(\pm 0.03)$ & $\mathbf{94.41}~(\pm 0.05)$ & $\mathbf{53.93}~(\pm 1.57)$  &  $\mathbf{38.76}~(\pm 0.52)$   \\\hline 
\hetlo, $\gamma = 0.95$ &  $89.62~(\pm 1.33)$  & $83.19~(\pm 1.70)$ & $71.10~(\pm 1.39)$ & $46.39~(\pm 0.87)$    \\ \hline 
\hetlo, $\gamma = 0.85$ &   $60.31~(\pm 3.04)$ &  $53.28~(\pm 2.47)$ & $120.72~(\pm 10.93)$  &   $59.67~(\pm 1.98)$   \\
\bottomrule   \vspace{-1.5em}
\end{tabular}
\end{table*}

\begin{figure}[!t]
\centering
\begin{subfigure}{0.49\textwidth}
\centering
\includegraphics[width=1\textwidth]{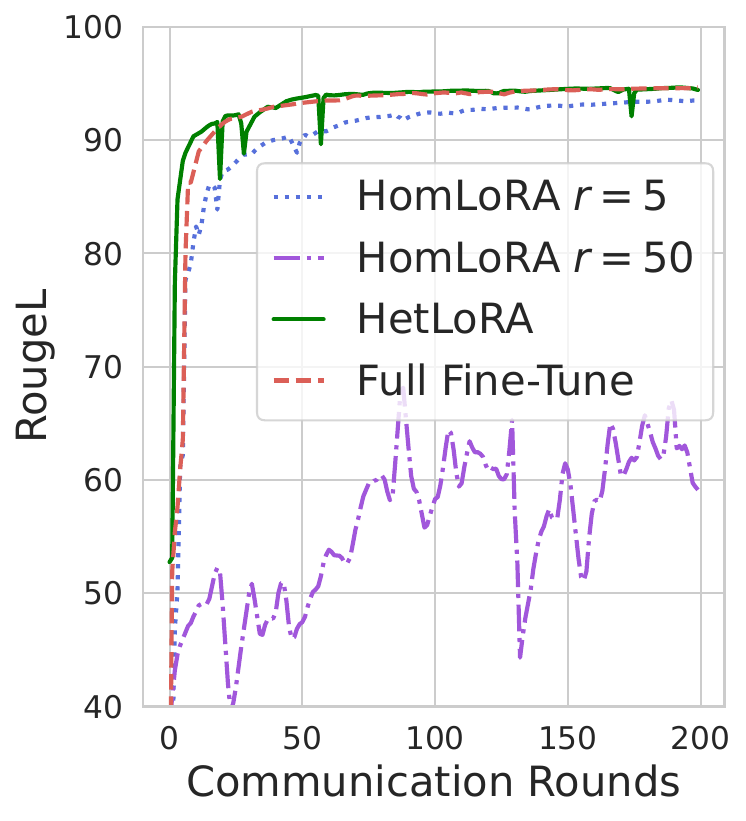}\vspace{-0.5em} \caption{Reddit}\vspace{-1em}
\end{subfigure}
\centering
\begin{subfigure}{0.49\textwidth}
\centering
\includegraphics[width=1\textwidth]{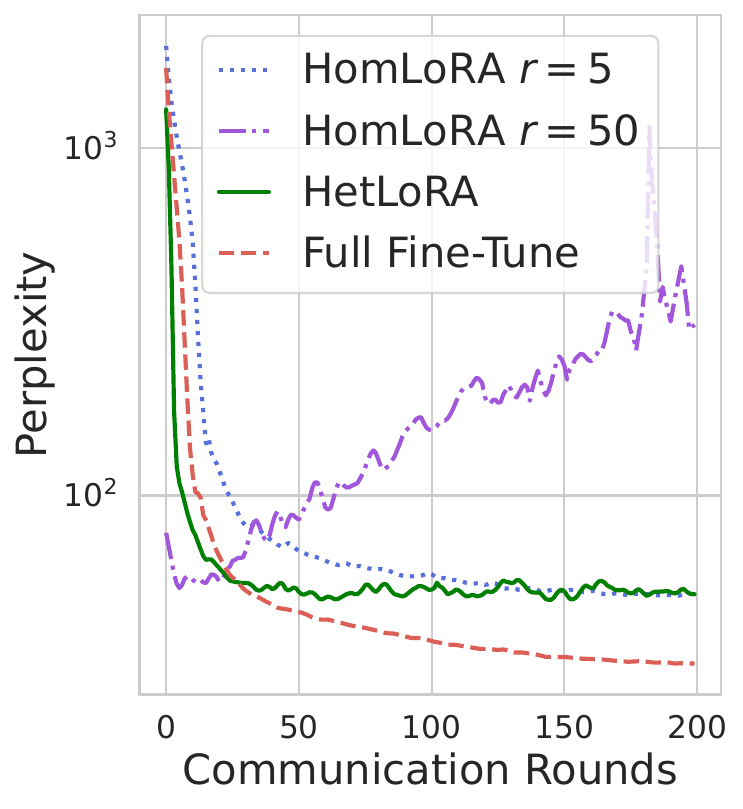}\vspace{-0.5em} \caption{MSC}\vspace{-1em}
\end{subfigure} 
    \caption{Comparison of the performance across homogeneous LoRA, heterogeneous LoRA, and full fine-tuning. Heterogeneous LoRA achieves better performance than homogeneous LoRA with fewer number of communication rounds. }
\label{fig:final_comp} 
\end{figure}

\begin{figure}
    \centering
    \includegraphics[width=0.9\textwidth]{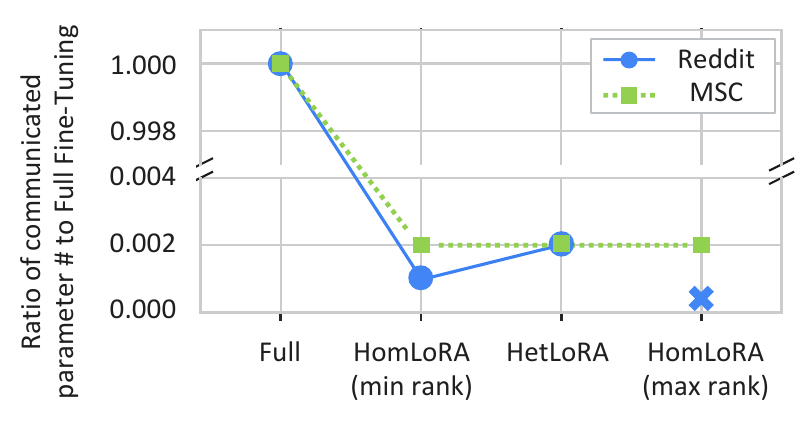}\vspace{-1em}
    \caption{Ratio of communicated number of parameters for different PEFT methods to full fine-tuning to achieve the target value for the metric where it is RougeL 80 for Reddit text summarization task and perplexity 150 for the multi-session chat response task. The `X' means that the target metric is not achieved even after convergence.\vspace{2em}}
    \label{fig:comp_comm}
\end{figure}

\textbf{Heterogeneous LoRA compared to Baselines.} Finally, we compare our proposed \hetlo with other baselines in \Cref{tab:final} and \Cref{fig:final_comp}. We see that \hetlo with $r_{\text{min}}=5$ and $r_{\text{max}}=50$ achieves faster training as well as better performance than homogeneous LoRA cases with both edge cases of the ranks $r\in\{5,50\}$ and reconostruction+SVD which was explained in \Cref{subsec:recon}. This implies that \hetlo is not only practical in the sense that clients are allowed to have their own rank values, it can also outperform the limited case of homogeneous LoRA where all clients have $r=\rmin$ or the impractical case where all clients have $r=\rmax$. We also observe that \hetlo achieves slightly lower performance than full fine-tuning. However, as shown in the blue text in \Cref{tab:final} that shows the number of trained parameters compared to the full fine-tuning case, full fine-tuning requires to train a much larger number of parameters compared to \hetlo, making it infeasible to train with ODFMs in practice. We also show in \Cref{fig:comp_comm} that to achieve the targeted performance for both Reddit and MSC task, \hetlo~requires significantly less number of parameters to be trained and communicated compared to full fine-tuning. Although for Reddit, \homlo has a slightly less number of parameters to be trained, the final achieved RougeL is outperformed by \hetlo as shown in \Cref{tab:final}.

\textbf{Effect of the Decaying Factor $\gamma$.} Lastly, we conduct an ablation study on the effect of the decaying factor $\gamma$ of \hetlo's local training step with self-rank pruning in \Cref{tab:decay}. We observed that aggressive pruning hurts the performance where $\gamma=0.85$ shows the worse performance across the varying $\gamma$ values. On the other hand, no pruning at all ($\gamma=1$) underperforms the case when there is pruning ($\gamma=0.99$), showing that reducing the noise introduced by large rank clients which data complexity is actually not that high indeed improves the performance. \vspace{-0.5em}

\section{Discussions and Concluding Remarks} 
In our work, we investigated federated fine-tuning for ODFMs that cater to device system and data heterogeneity with our proposed \hetlo. We show that \hetlo is not only practical but also achieves better training speed, communication/computation efficiency, and final performance compared to homogeneous LoRA. We also show interesting findings consistent with previous literature~\cite{hu2021lora, zhang2023adalora} that increasing ranks does not always help for homogeneous LoRA. Our findings in this work opens up several questions worth investigating. For instance, if the settings allow us to assign specific ranks to clients what will be the effective way to assign the ranks across clients for better convergence and performance? Another important next step of our work includes pursuing the theoretical convergence and generalization of heterogeneous LoRA.

\section{Limitations}
In this work, we address tackling system and data heterogeneity in federated fine-tuning of on-device foundation models. Our work is motivated by clients being able to carry different ranks for the LoRA fine-tuning method depending on their available resources, and thus exploiting this characteristic to improve federated fine-tuning with heterogeneous LoRA. However, our work assumes that the rank distribution across clients (which is analogous to how system resources are distributed across clients) is independent to the data distribution. There can be scenarios in which this is not necessarily the case where the rank and data distribution can be correlated. For instance, more affluent populations can have better off devices with larger resource capacity, and may have data distributions different to that of less affluent populations. Such correlation should be explored for future work to better understand the implications of heterogenoeus LoRA. 
\bibliography{dist_sgd}

\begin{thebibliography}{46}
\expandafter\ifx\csname natexlab\endcsname\relax\def\natexlab#1{#1}\fi

\bibitem[{Babakniya et~al.(2023)Babakniya, Elkordy, Ezzeldin, Liu, Song, El-Khamy, and Avestimehr}]{baba2023slora}
Sara Babakniya, Ahmed~Roushdy Elkordy, Yahya~H. Ezzeldin, Qingfeng Liu, Kee-Bong Song, Mostafa El-Khamy, and Salman Avestimehr. 2023.
\newblock Slora: Federated parameter efficient fine-tuning of language models.
\newblock \emph{CoRR}, abs/2308.06522.

\bibitem[{Bommasani et~al.(2022)Bommasani, Hudson, and Ehsan~Adeli}]{rishi2022fm}
Rishi Bommasani, Drew~A. Hudson, and et.~al. Ehsan~Adeli. 2022.
\newblock On the opportunities and risks of foundation models.
\newblock \emph{arXiv preprint arXiv:2108.07258}.

\bibitem[{Bonawitz et~al.(2016)Bonawitz, Ivanov, Kreuter, Marcedone, McMahan, Patel, Ramage, Segal, and Seth}]{bonawitz2016practical}
Keith Bonawitz, Vladimir Ivanov, Ben Kreuter, Antonio Marcedone, H.~Brendan McMahan, Sarvar Patel, Daniel Ramage, Aaron Segal, and Karn Seth. 2016.
\newblock Practical secure aggregation for federated learning on user-held data.
\newblock In \emph{NIPS Workshop on Private Multi-Party Machine Learning}.

\bibitem[{Brown et~al.(2020)Brown, Mann, Ryder, Subbiah, Kaplan, Dhariwal, Neelakantan, Shyam, Sastry, Askell, Agarwal, Herbert-Voss, Krueger, Henighan, Child, Ramesh, Ziegler, Wu, Winter, Hesse, Chen, Sigler, Litwin, Gray, Chess, Clark, Berner, McCandlish, Radford, Sutskever, and Amodei}]{brown2020gpt3}
Tom~B. Brown, Benjamin Mann, Nick Ryder, Melanie Subbiah, Jared Kaplan, Prafulla Dhariwal, Arvind Neelakantan, Pranav Shyam, Girish Sastry, Amanda Askell, Sandhini Agarwal, Ariel Herbert-Voss, Gretchen Krueger, Tom Henighan, Rewon Child, Aditya Ramesh, Daniel~M. Ziegler, Jeffrey Wu, Clemens Winter, Christopher Hesse, Mark Chen, Eric Sigler, Mateusz Litwin, Scott Gray, Benjamin Chess, Jack Clark, Christopher Berner, Sam McCandlish, Alec Radford, Ilya Sutskever, and Dario Amodei. 2020.
\newblock Language models are few-shot learners.
\newblock \emph{arXiv preprint arXiv:2005.14165}.

\bibitem[{Bu et~al.(2022)Bu, Wang, Zha, and Karypis}]{bu2022differentially}
Zhiqi Bu, Yu-Xiang Wang, Sheng Zha, and George Karypis. 2022.
\newblock Differentially private bias-term only fine-tuning of foundation models.
\newblock \emph{arXiv preprint arXiv:2210.00036}.

\bibitem[{Chen et~al.(2022)Chen, Xu, Guo, Wang, Zhang, and Wang}]{chen2022fedtune}
Jinyu Chen, Wenchao Xu, Song Guo, Junxiao Wang, Jie Zhang, and Haozhao Wang. 2022.
\newblock Fedtune: A deep dive into efficient federated fine-tuning with pre-trained transformers.
\newblock \emph{CoRR}, abs/2211.08025.

\bibitem[{Devlin et~al.(2019)Devlin, Chang, Lee, and Toutanova}]{devlin2019bert}
Jacob Devlin, Ming-Wei Chang, Kenton Lee, and Kristina Toutanova. 2019.
\newblock Bert: Pre-training of deep bidirectional transformers for language understanding.
\newblock \emph{arXiv preprint arXiv:1810.04805}.

\bibitem[{Driess et~al.(2023)Driess, Xia, Sajjadi, Lynch, Chowdhery, Ichter, Wahid, Tompson, Vuong, Yu, Huang, Chebotar, Sermanet, Duckworth, Levine, Vanhoucke, Hausman, Toussaint, Greff, Zeng, Mordatch, and Florence}]{driess2023palme}
Danny Driess, Fei Xia, Mehdi S.~M. Sajjadi, Corey Lynch, Aakanksha Chowdhery, Brian Ichter, Ayzaan Wahid, Jonathan Tompson, Quan Vuong, Tianhe Yu, Wenlong Huang, Yevgen Chebotar, Pierre Sermanet, Daniel Duckworth, Sergey Levine, Vincent Vanhoucke, Karol Hausman, Marc Toussaint, Klaus Greff, Andy Zeng, Igor Mordatch, and Pete Florence. 2023.
\newblock Palm-e: An embodied multimodal language model.
\newblock \emph{arXiv preprint arXiv:2303.03378}.

\bibitem[{Google(2022)}]{chowd2022palm}
Google. 2022.
\newblock Palm: Scaling language modeling with pathways.
\newblock \emph{arXiv preprint arXiv:2204.02311}.

\bibitem[{Google(2023)}]{google2023palm2}
Google. 2023.
\newblock Palm 2 technical report.
\newblock \emph{arXiv preprint arXiv:2305.1040}.

\bibitem[{{Google DeepMind}(2023)}]{google_palm2}
{Google DeepMind}. 2023.
\newblock Introducing palm2.
\newblock \url{https://blog.google/technology/ai/google-palm-2-ai-large-language-model/}.

\bibitem[{Guo et~al.(2023)Guo, Guo, and Wang}]{tao2023fmper}
Tao Guo, Song Guo, and Junxiao Wang. 2023.
\newblock \href {https://doi.org/10.1145/3543507.3583518} {Pfedprompt: Learning personalized prompt for vision-language models in federated learning}.
\newblock In \emph{Proceedings of the ACM Web Conference 2023}, WWW '23, page 1364–1374, New York, NY, USA. Association for Computing Machinery.

\bibitem[{Guo et~al.(2022)Guo, Guo, Wang, and Xu}]{guo2022promptfl}
Tao Guo, Song Guo, Junxiao Wang, and Wenchao Xu. 2022.
\newblock Promptfl: Let federated participants cooperatively learn prompts instead of models — federated learning in age of foundation model.
\newblock \emph{CoRR}, abs/2208.11625.

\bibitem[{Guruswami and Kannan(2012)}]{guru2012lecnote}
Venkatesan Guruswami and Ravi Kannan. 2012.
\newblock Lecture notes in computer science theory for the information age.

\bibitem[{Houlsby et~al.(2019)Houlsby, Giurgiu, Jastrzebski, Morrone, de~Laroussilhe, Gesmundo, Attariyan, and Gelly}]{houlsby2019adapter}
Neil Houlsby, Andrei Giurgiu, Stanisław Jastrzebski, Bruna Morrone, Quentin de~Laroussilhe, Andrea Gesmundo, Mona Attariyan, and Sylvain Gelly. 2019.
\newblock Parameter-efficient transfer learning for nlp.
\newblock In \emph{Proceedings of the International Conference on Machine Learning (ICML)}.

\bibitem[{Hu et~al.(2021)Hu, Shen, Wallis, Allen-Zhu, Li, Wang, Wang, and Chen}]{hu2021lora}
Edward~J. Hu, Yelong Shen, Phillip Wallis, Zeyuan Allen-Zhu, Yuanzhi Li, Shean Wang, Lu~Wang, and Weizhu Chen. 2021.
\newblock Lora: Low-rank adaptation of large language models.
\newblock In \emph{International Conference on Learning Representations (ICLR)}.

\bibitem[{Kojima et~al.(2022)Kojima, Gu, Reid, Matsuo, and Iwasawa}]{kojima2022fmzeroshot}
Takeshi Kojima, Shixiang~Shane Gu, Machel Reid, Yutaka Matsuo, and Yusuke Iwasawa. 2022.
\newblock Large language models are zero-shot reasoners.
\newblock In \emph{The 36th Conference on Neural Information Processing Systems (NeurIPS 2022)}.

\bibitem[{Lee et~al.(2019)Lee, Tang, and Lin}]{lee2019would}
Jaejun Lee, Raphael Tang, and Jimmy Lin. 2019.
\newblock What would elsa do? freezing layers during transformer fine-tuning.
\newblock \emph{arXiv preprint arXiv:1911.03090}.

\bibitem[{Legate et~al.(2023)Legate, Bernier, Caccia, Oyallon, and Belilovsky}]{gwen2023lastlayerfm}
Gwen Legate, Nicolas Bernier, Lucas Caccia, Edouard Oyallon, and Eugene Belilovsky. 2023.
\newblock Guiding the last layer in federated learning with pre-trained models.
\newblock In \emph{Workshop of Federated Learning and Analytics in Practice@ICML}.

\bibitem[{Lester et~al.(2021)Lester, Al-Rfou, and Constant}]{lester2021prompt}
Brian Lester, Rami Al-Rfou, and Noah Constant. 2021.
\newblock The power of scale for parameter-efficient prompt tuning.
\newblock In \emph{Empirical Methods in Natural Language Processing (EMNLP)}.

\bibitem[{Li and Liang(2021)}]{li2021prefix}
Xiang~Lisa Li and Percy Liang. 2021.
\newblock Prefix-tuning: Optimizing continuous prompts for generation.
\newblock \emph{arXiv preprint arXiv:2101.00190}.

\bibitem[{Lin(2004)}]{lin2004rouge}
Chin-Yew Lin. 2004.
\newblock \href {https://aclanthology.org/W04-1013} {{ROUGE}: A package for automatic evaluation of summaries}.
\newblock In \emph{Text Summarization Branches Out}, pages 74--81, Barcelona, Spain. Association for Computational Linguistics.

\bibitem[{Manoel et~al.(2023)Manoel, del Carmen Hipolito~Garcia, Baumel, Su, Chen, Sim, Miller, Karmon, and Dimitriadis}]{manoel2023fedmed}
Andre Manoel, Mirian del Carmen Hipolito~Garcia, Tal Baumel, Shize Su, Jialei Chen, Robert Sim, Dan Miller, Danny Karmon, and Dimitrios Dimitriadis. 2023.
\newblock Federated multilingual models for medical transcript analysis.
\newblock In \emph{Conference on Health, Inference, and Learning (CHIL)}, pages 147--162.

\bibitem[{McMahan et~al.(2017)McMahan, Moore, Ramage, Hampson, and y~Arcas}]{mcmahan2017communication}
H.~Brendan McMahan, Eider Moore, Daniel Ramage, Seth Hampson, and Blaise~Ag\o{u}ra y~Arcas. 2017.
\newblock \href {https://arxiv.org/abs/1602.05629} {{Communication-Efficient Learning of Deep Networks from Decentralized Data}}.
\newblock \emph{International Conference on Artificial Intelligenece and Statistics (AISTATS)}.

\bibitem[{OpenAI(2023)}]{openai2023gpt4}
OpenAI. 2023.
\newblock Gpt-4 technical report.
\newblock \emph{arXiv preprint arXiv:submit/4812508}.

\bibitem[{Radford et~al.(2021)Radford, Kim, Hallacy, Ramesh, Goh, Agarwal, Sastry, Askell, Mishkin, Clark, Krueger, and Sutskever}]{radford2021clip}
Alec Radford, Jong~Wook Kim, Chris Hallacy, Aditya Ramesh, Gabriel Goh, Sandhini Agarwal, Girish Sastry, Amanda Askell, Pamela Mishkin, Jack Clark, Gretchen Krueger, and Ilya Sutskever. 2021.
\newblock Learning transferable visual models from natural language supervision.
\newblock \emph{arXiv preprint arXiv:2103.00020}.

\bibitem[{Sahu et~al.(2020)Sahu, Li, Sanjabi, Zaheer, Talwalkar, and Smith}]{sahu2019federated}
Anit~Kumar Sahu, Tian Li, Maziar Sanjabi, Manzil Zaheer, Ameet Talwalkar, and Virginia Smith. 2020.
\newblock Federated optimization for heterogeneous networks.
\newblock In \emph{Proceedings of the 3rd MLSys Conference}.

\bibitem[{Sedoc et~al.(2019)Sedoc, Ippolito, Kirubarajan, Thirani, Ungar, and Callison-Burch}]{sedoc2019chateval}
Joao Sedoc, Daphne Ippolito, Arun Kirubarajan, Jai Thirani, Lyle Ungar, and Chris Callison-Burch. 2019.
\newblock Chateval: A tool for chatbot evaluation.
\newblock \emph{Proceedings of NAACL-HLT}.

\bibitem[{Shoham and Rappoport(2023)}]{shoham2023medfedbert}
Ofir~Ben Shoham and Nadav Rappoport. 2023.
\newblock Federated learning of medical concepts embedding using behrt.
\newblock \emph{arXiv preprint arXiv:2305.13052}.

\bibitem[{Shysheya et~al.(2023)Shysheya, Bronskill, Patacchiola, Nowozin, and Turner}]{shy2023peftfit}
Aliaksandra Shysheya, John~F Bronskill, Massimiliano Patacchiola, Sebastian Nowozin, and Richard~E Turner. 2023.
\newblock Fit: Parameter efficient few-shot transfer learning for personalized and federated image classification.
\newblock \emph{International Conference on Learning Representations (ICLR)}.

\bibitem[{Touvron et~al.(2023)Touvron, Martin, Stone, Albert, Almahairi, Babaei, Bashlykov, Batra, Bhargava, Bhosale et~al.}]{touvron2023llama}
Hugo Touvron, Louis Martin, Kevin Stone, Peter Albert, Amjad Almahairi, Yasmine Babaei, Nikolay Bashlykov, Soumya Batra, Prajjwal Bhargava, Shruti Bhosale, et~al. 2023.
\newblock Llama 2: Open foundation and fine-tuned chat models.
\newblock \emph{arXiv preprint arXiv:2307.09288}.

\bibitem[{V{\"o}lske et~al.(2017)V{\"o}lske, Potthast, Syed, and Stein}]{volske2017reddit}
Michael V{\"o}lske, Martin Potthast, Shahbaz Syed, and Benno Stein. 2017.
\newblock \href {https://doi.org/10.18653/v1/W17-4508} {{TL};{DR}: Mining {R}eddit to learn automatic summarization}.
\newblock In \emph{Proceedings of the Workshop on New Frontiers in Summarization}, pages 59--63, Copenhagen, Denmark. Association for Computational Linguistics.

\bibitem[{Wang et~al.(2021)Wang, Charles, Xu, Joshi, McMahan, Al-Shedivat, Andrew, Avestimehr, Daly, Data et~al.}]{wang2021field}
Jianyu Wang, Zachary Charles, Zheng Xu, Gauri Joshi, H~Brendan McMahan, Maruan Al-Shedivat, Galen Andrew, Salman Avestimehr, Katharine Daly, Deepesh Data, et~al. 2021.
\newblock A field guide to federated optimization.
\newblock \emph{arXiv preprint arXiv:2107.06917}.

\bibitem[{Wang et~al.(2019)Wang, Tuor, Salonidis, Leung, Makaya, He, and Chan}]{wang2019systemc}
Shiqiang Wang, Tiffany Tuor, Theodoros Salonidis, Kin~K. Leung, Christian Makaya, Ting He, and Kevin Chan. 2019.
\newblock Adaptive federated learning in resource constrained edge computing systems.
\newblock \emph{IEEE Journal on Selected Areas in Communications}, 37(6):1205 -- 1221.

\bibitem[{Wei et~al.(2021)Wei, Xie, and Ma}]{wei2021pretrained}
Colin Wei, Sang~Michael Xie, and Tengyu Ma. 2021.
\newblock Why do pretrained language models help in downstream tasks? an analysis of head and prompt tuning.
\newblock \emph{Advances in Neural Information Processing Systems}, 34:16158--16170.

\bibitem[{Wortsman et~al.(2023)Wortsman, Gururangan, Li, Farhadi, Schmidt, Rabbat, and Morcos}]{wortsman2023distributedlofi}
Mitchell Wortsman, Suchin Gururangan, Shen Li, Ali Farhadi, Ludwig Schmidt, Michael Rabbat, and Ari~S. Morcos. 2023.
\newblock lo-fi: distributed fine-tuning without communication.
\newblock \emph{Transactions on Machine Learning Research (TMLR)}.

\bibitem[{Xu et~al.(2021)Xu, Szlam, and Weston}]{xu2021msc}
Jing Xu, Arthur Szlam, and Jason Weston. 2021.
\newblock Beyond goldfish memory: Long-term open-domain conversation.
\newblock \emph{arXiv preprint arXiv:2107.07567}.

\bibitem[{Yi et~al.(2023)Yi, Yu, Wang, and Liu}]{yi2023fedlora}
Liping Yi, Han Yu, Gang Wang, and Xiaoguang Liu. 2023.
\newblock Fedlora: Model-heterogeneous personalized federated learning with lora tuning.
\newblock \emph{arXiv preprint arXiv:2310.13283}.

\bibitem[{Yu et~al.(2023)Yu, Muñoz, and Jannesari}]{yu2023ffm}
Sixing Yu, J.~Pablo Muñoz, and Ali Jannesari. 2023.
\newblock Federated foundation models: Privacy-preserving and collaborative learning for large models.
\newblock \emph{arXiv preprint arXiv:2305.11414}.

\bibitem[{Zaken et~al.(2021)Zaken, Ravfogel, and Goldberg}]{zaken2021bitfit}
Elad~Ben Zaken, Shauli Ravfogel, and Yoav Goldberg. 2021.
\newblock Bitfit: Simple parameter-efficient fine-tuning for transformer-based masked language-models.
\newblock \emph{arXiv preprint arXiv:2106.10199}.

\bibitem[{Zhang et~al.(2023{\natexlab{a}})Zhang, Vahidian, Kuo, Li, Zhang, and Chen}]{jianyi2022fedgpt}
Jianyi Zhang, Saeed Vahidian, Martin Kuo, Chunyuan Li, Ruiyi Zhang, and Guoyin Wangand~Yiran Chen. 2023{\natexlab{a}}.
\newblock Towards building the federated gpt: Federated instruction tuning.
\newblock \emph{CoRR}, abs/2305.05644.

\bibitem[{Zhang et~al.(2023{\natexlab{b}})Zhang, Chen, Bukharin, He, Cheng, Chen, and Zhao}]{zhang2023adalora}
Qingru Zhang, Minshuo Chen, Alexander Bukharin, Pengcheng He, Yu~Cheng, Weizhu Chen, and Tuo Zhao. 2023{\natexlab{b}}.
\newblock Adaptive budget allocation for parameter-efficient fine-tuning.
\newblock In \emph{The 11th International Conference on Learning Representations (ICLR)}.

\bibitem[{Zhang et~al.(2018)Zhang, Dinan, Urbanek, Szlam, Kiela, and Weston}]{zhang2018perplexity}
Saizheng Zhang, Emily Dinan, Jack Urbanek, Arthur Szlam, Douwe Kiela, and Jason Weston. 2018.
\newblock Personalizing dialogue agents: I have a dog, do you have pets too?
\newblock \emph{In Proceedings of the 56th Annual Meeting of the Association for Computational Linguistics (ACL)}.

\bibitem[{Zhang et~al.(2023{\natexlab{c}})Zhang, Hu, Zhang, Zhang, Wang, Qu, and Xu}]{zhang2023fedlegal}
Zhuo Zhang, Xiangjing Hu, Jingyuan Zhang, Yating Zhang, Hui Wang, Lizhen Qu, and Zenglin Xu. 2023{\natexlab{c}}.
\newblock Fedlegal: The first real-world federated learning benchmark for legal nlp.
\newblock \emph{In Proceedings of the 61st Annual Meeting of the Association for Computational Linguistics (ACL)}.

\bibitem[{Zhang et~al.(2023{\natexlab{d}})Zhang, Yang, Dai, Wang, Yu, Qu, and Xu}]{zhang2023pretrainfinetune}
Zhuo Zhang, Yuanhang Yang, Yong Dai, Qifan Wang, Yue Yu, Lizhen Qu, and Zenglin Xu. 2023{\natexlab{d}}.
\newblock Fedpetuning: When federated learning meets the parameter-efficient tuning methods of pre-trained language models.
\newblock \emph{Findings of the Association for Computational Linguistics (ACL)}.

\bibitem[{Zhou et~al.(2023)Zhou, Li, Li, Yu, Liu, Wang, Zhang, Ji, Yan, He, Peng, Li, Wu, Liu, Xie, Xiong, Pei, Yu, and Sun}]{zhou2023fmsurvey}
Ce~Zhou, Qian Li, Chen Li, Jun Yu, Yixin Liu, Guangjing Wang, Kai Zhang, Cheng Ji, Qiben Yan, Lifang He, Hao Peng, Jianxin Li, Jia Wu, Ziwei Liu, Pengtao Xie, Caiming Xiong, Jian Pei, Philip~S. Yu, and Lichao Sun. 2023.
\newblock A comprehensive survey on pretrained foundation models: A history from bert to chatgpt.
\newblock \emph{arXiv preprint arXiv:2302.09419}.

\end{thebibliography}

\clearpage
\newpage
\appendix

\comment{
\section{Additional Experimental Results}\label{sec:aer}
\begin{figure}[!htb]
\begin{subfigure}{0.49\textwidth}
    \includegraphics[width=1\textwidth]{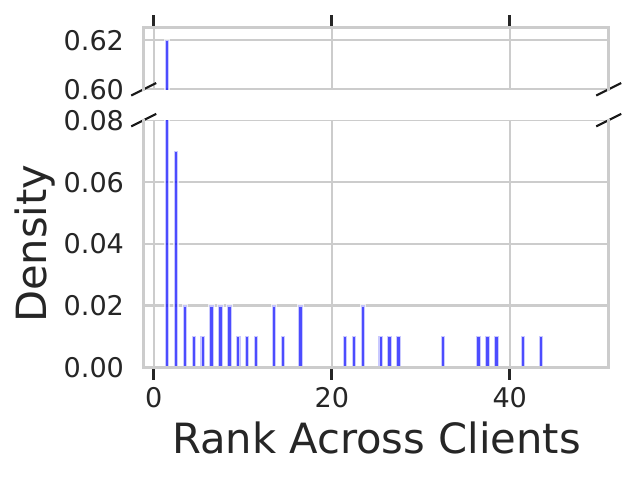} \vspace{-2em}
\caption{}
\end{subfigure}
\begin{subfigure}{0.49\textwidth}
    \includegraphics[width=1\textwidth]{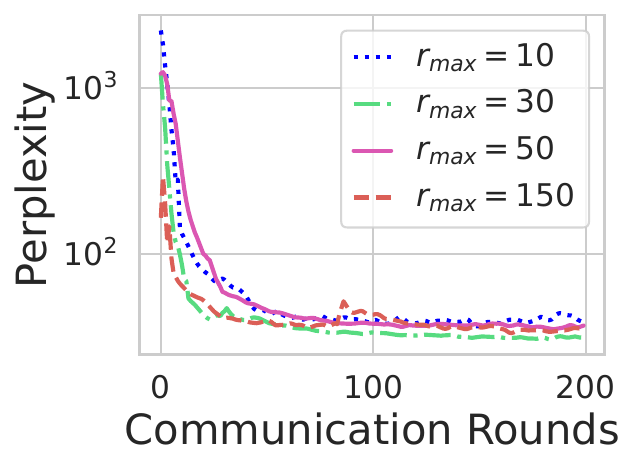} 
    \caption{}
\end{subfigure}
\vspace{-2em}
    \caption{\small Performance of heterogeneous LoRA with varying $r_{\text{max}}$. The best performing case is $r_{\text{max}}=30,r_{\text{min}}=1$ showing that increasing the maximum rank does not always help. It is also observable that for $r_{\text{min}}=5$ the discrepancies across maximum ranks are nearly negligible due to overfitting showing that the minimum rank $r_{\text{min}}$ plays an important role in preventing overfitting.}
\label{fig:hetlora_max}
\end{figure}

\begin{table}[!htb] \centering 
\small
\caption{Ablation study on fixing the left module for \hetlo}
\begin{tabular}{@{}lcc@{}}\toprule
& PaLM 2-XXS & PaLM 2-XS  \\\hline
Reddit & $94.42 (\pm 0.13)$ & $94.23 (\pm 0.39)$  \\  \hline
MSC & $55.65 (\pm 0.09)$ & $35.58 (\pm 0.68)$   \\
\bottomrule \label{tab:fair} \vspace{-2em}
\end{tabular}
\end{table}

}

\end{document}